\documentclass[%
reprint,
nofootinbib,
 amsmath,amssymb,
aps,
superscriptaddress,
showkeys,
longbibliography
]{revtex4-1}

\usepackage{enumitem}

\usepackage{xr}
\usepackage{tabularx}
\usepackage{ltablex}
\usepackage{booktabs}
\usepackage{graphicx}
\usepackage{dcolumn}
\usepackage{bm}
\usepackage{microtype}
\usepackage{gensymb}
\usepackage{url}
\usepackage[breaklinks, hidelinks, colorlinks=true,linkcolor=blue,citecolor=black,hypertexnames=false]{hyperref}
\usepackage{color, colortbl}
\usepackage[table,xcdraw]{xcolor}

\makeatletter
\renewcommand\frontmatter@abstractwidth{\dimexpr0.9\textwidth\relax}
\makeatother

\usepackage{algorithm}
\usepackage{algpseudocode}
\usepackage{multirow}

\makeatletter
\newcommand*{\addFileDependency}[1]{
  \typeout{(#1)}
  \@addtofilelist{#1}
  \IfFileExists{#1}{}{\typeout{No file #1.}}
}
\makeatother

\makeatletter
\renewcommand\subparagraph{\@startsection{subparagraph}{5}{\parindent}%
    {3.25ex \@plus1ex \@minus .2ex}%
    {-1em}%
    {\normalfont\normalsize\bfseries}}
\makeatother

\begin{document}
\preprint{1}

\title{VessShape: Few-shot 2D blood vessel segmentation by leveraging shape priors from synthetic images}

\author{Wesley N. Galvão}
\email[]{wesleygalvao@estudante.ufscar.br}
\affiliation{Department of Computer Science, Federal University of S\~ao Carlos, S\~ao Carlos, SP, Brazil}

\author{Cesar H. Comin}
\email[Corresponding author: ]{comin@ufscar.br}
\affiliation{Department of Computer Science, Federal University of S\~ao Carlos, S\~ao Carlos, SP, Brazil}

\date{\today}

\begin{abstract}

Semantic segmentation of blood vessels is an important task in medical image analysis, but its progress is often hindered by the scarcity of large annotated datasets and the poor generalization of models across different imaging modalities. A key aspect is the tendency of Convolutional Neural Networks (CNNs) to learn texture-based features, which limits their performance when applied to new domains with different visual characteristics. We hypothesize that leveraging geometric priors of vessel shapes, such as their tubular and branching nature, can lead to more robust and data-efficient models. To investigate this, we introduce VessShape, a methodology for generating large-scale 2D synthetic datasets designed to instill a shape bias in segmentation models. VessShape images contain procedurally generated tubular geometries combined with a wide variety of foreground and background textures, encouraging models to learn shape cues rather than textures. We demonstrate that a model pre-trained on VessShape images achieves strong few-shot segmentation performance on two real-world datasets from different domains, requiring only four to ten samples for fine-tuning. Furthermore, the model exhibits notable zero-shot capabilities, effectively segmenting vessels in unseen domains without any target-specific training. Our results indicate that pre-training with a strong shape bias can be an effective strategy to overcome data scarcity and improve model generalization in blood vessel segmentation.

\end{abstract}


\maketitle
\thispagestyle{plain}

\section{Introduction}
\label{sec:introduction}

Semantic segmentation of blood vessels is an active area of research, driven by the demand for precise and automated analysis of medical images in both human and animal tissues. A significant challenge in this task is the labor-intensive process of manual annotation, which requires domain expertise to create accurate segmentation masks. This annotation bottleneck has led to a scarcity of large-scale datasets, which limits both the training of deep learning models and the development of new methods. For instance, widely used public datasets such as DRIVE \cite{Staal2004} and CHASE\_DB1 \cite{Fraz2012ensemble} contain only a few dozen annotated images each. Although recent efforts have introduced new datasets \cite{jin2022fives, fhima2024lunet}, the availability of large and diverse collections remains limited. This problem is compounded by significant domain shifts between different imaging modalities, such as retinal fundus photography and cerebral cortex microscopy. Variations in texture, vessel density, and caliber hinder the ability of models trained in one domain to generalize to another.

Transfer learning is a common strategy to address these limitations. By reusing representations learned on a source domain, models can achieve better performance on a target domain, even with limited data. Techniques such as fine-tuning and domain adaptation allow models pre-trained on large-scale natural image datasets, like ImageNet \cite{JiaDeng2009}, to be adapted for biomedical segmentation tasks \cite{zoetmulderDomainTaskspecificTransfer2022}.

However, a potential pitfall of standard transfer learning is the inherent bias of Convolutional Neural Networks (CNNs) toward learning texture-based features rather than geometric shapes \cite{geirhos2018, islam2021shape}. This texture bias can limit generalization across domains with different visual styles. Research has shown that models trained with an emphasis on shape cues exhibit improved performance and robustness \cite{geirhos2018}. This suggests that transferring shape representations, rather than texture-rich features, could be a more effective strategy for few-shot blood vessel segmentation.

\begin{figure}[tbp]
    \centering
    \includegraphics[width=\columnwidth]{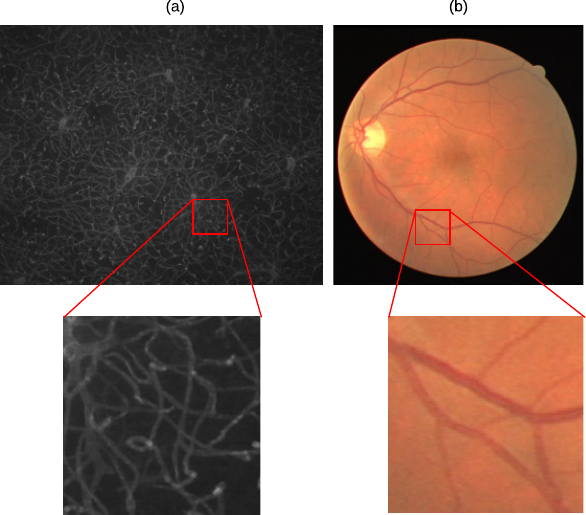}
    \caption{Illustration of the universality of vessel shape. (a) Fluorescence microscopy sample of a mouse cortex. (b) Fundus photograph of a human eye. Despite having different textures, the vessels have similar shapes.}
    \label{f:motivation}
\end{figure}

The task of blood vessel segmentation is particularly well suited for a shape-centric approach. The fundamental geometry of blood vessels is a consistent prior across diverse imaging modalities, from retinal fundus photographs to cerebral cortex microscopy images. While textures and other visual characteristics may vary significantly between these domains, the underlying shape remains a universal identifier. For example, a human observer who has learned to identify vessels in one modality can readily recognize them in another, as illustrated in Figure~\ref{f:motivation}.

Based on this observation, we hypothesize that a model pre-trained on a source domain with a strong shape bias will require only a few annotated samples to adapt to a new target domain. We expect such a model to outperform a model trained from scratch on the target domain by effectively reusing its learned geometric priors. To test this hypothesis, we introduce VessShape, a methodology for creating synthetic datasets designed for pre-training shape-aware vessel segmentation models. VessShape consists of 2D images with tubular, vessel-like structures paired with a wide variety of foreground and background textures. By fixing the geometric priors while diversifying the textures, the dataset explicitly encourages models to learn robust shape features over superficial texture cues.

We demonstrate that a model pre-trained on VessShape images can accurately segment vasculature in two different target domains using only four to ten annotated samples for fine-tuning. Furthermore, we observe remarkable zero-shot capabilities, where the pre-trained model can segment vessels in new domains without any target-specific training.

\section{Related Works}
\label{sec:related}

Many previous studies have considered shape priors for medical image segmentation~\cite{bohlender2021survey,heimann2009statistical,cootes1995active}. This is possible when organelles, cells, or organs have a known shape and the segmentation process can assume an optimal shape or additional optimization criteria such as the requirement of smooth borders. Instance segmentation is probably the most common task in which shape priors have been explored. A particular challenge with blood vessel identification is that it usually requires a semantic segmentation of the image. 

Before the popularity of deep learning methods, local shape priors were the dominant approach to vessel segmentation. Blood vessels tend to have a tubular structure. Thus, the usual approach was to develop filters aimed at identifying tubular objects. A popular line of research was based on the eigenvalues of the Hessian matrix~\cite{fraz2012blood,sato1998three}. Arguably, the most popular Hessian-based method is the Frangi filter~\cite{frangi1998multiscale}, which consists of combining the eigenvalues to define a tubularity score for the pixels of the vessels. Another popular approach was based on the definition of line or Gabor filter templates to identify relevant vessel structures. An important drawback of these methods is that the tubularity assumption is not valid at bifurcation and termination points.

With the emergence of deep learning, some works have explored adding shape priors during network training~\cite{bohlender2021survey}. Priors have been added on the network input using vesselness filters~\cite{affane2022robust,hu2024domain,garret2024deep}, on the network architecture using learnable vesselness or Gabor filters~\cite{chen2023learnable,fu2018frangi,volkov2025modification}, and on the network output using topology-aware loss functions~\cite{shit2021cldice,hu2019topology,berger2024topologically}. For neural networks, a simpler and likely more flexible alternative is a data-based approach that focuses on generating a large and diverse set of images containing a priori information about the structure of blood vessels. Our approach differs from these methods by using synthetic data to explicitly instill a strong shape bias while systematically diversifying textures. The closest approach is to generate synthetic images that are as similar as possible to the samples in the dataset. This has been done using two main strategies: i) creating an appearance model of the vessels and image background~\cite{tetteh2020deepvesselnet,wittmann2025vesselfm,wittmann2024simulation,mathys2025synthetic} and ii) using generative models to synthesize new samples from real images.

Modeling-based approaches usually start by generating a biologically plausible topology of the vasculature, followed by the definition of varying radii for vessel segments and the inclusion of texture for the vessels and the background. The typical noise found in the imaging modality of interest is also modeled. An important recent work in this direction is a foundation model called VesselFM~\cite{wittmann2025vesselfm} that was trained on a large number of synthetic and real images.

Regarding generative models, most works in the literature use Generative Adversarial Networks (GANs) to create samples~\cite{you2022application,andreini2021two,tavakkoli2020novel,costa2017end}. Synthetic samples conform to the learned patterns from the real dataset, allowing the generation of realistic images. Recent works considered diffusion models for the same task~\cite{go2024generation,guo2025vesseldiffusion,wang2025vastsd}. Some studies also developed style transfer approaches for domain adaptation between different datasets~\cite{peng2022unsupervised,chen2023segmentation,chen2021real}.

The main drawback of the aforementioned works is that the network is trained to reproduce the shape and texture of blood vessels in specific datasets. Changes in the texture of the vessels due to diseases or modifications in the imaging device can lead to low segmentation performance. In addition, models must be trained for specific imaging modalities, even if annotated data are scarce. Our approach aims at training neural networks to segment any vascular tissue that follows the shape priors acquired from VessShape.

\section{Methodology}
\label{s:methodology}

\subsection{The VessShape Generator}

The geometry of the synthetic images in VessShape\footnote{{Code repository}: \url{https://github.com/galvaowesley/vess-shape-dataset}} is defined using Bézier curves, which allow a flexible and controlled representation of tubular shapes. Each vascular segment is described by a $n$th-order Bézier curve with control points $\{\mathbf{p}_i\}_{i=0}^n$. Segment tortuosity is adjusted by small perturbations to these control points, ensuring a realistic and diverse vessel geometry. The Bézier curve $\mathbf{c}(t)$ of a vessel segment is given by Equation~\ref{eq:bezier}, where $t$ varies from 0 to 1.

\begin{equation}
\mathbf{c}(t) \,=\, \sum_{i=0}^{n} \binom{n}{i} (1-t)^{n-i} t^{i} \, \mathbf{p}_i,
\label{eq:bezier}
\end{equation}

To generate a curve, the first ($\mathbf{p}_0$) and last ($\mathbf{p}_n$) control points are sampled uniformly at random from the image domain. The remaining control points are generated by defining $n-1$ equally spaced points on the line connecting $\mathbf{p}_0$ and $\mathbf{p}_n$. Each of these points is then displaced by a random amount along a normal vector $\mathbf{n}_l$. This normal vector is a unit vector perpendicular to the line between $\mathbf{p}_0$ and $\mathbf{p}_n$. The displacement amount is drawn from a uniform distribution in $[-\delta,\delta]$. Lower values of $\delta$ lead to straighter curves.

For the generation of the binary mask $M$, each curve is discretized by sampling points at a sufficient resolution to capture its curvature, which are then sequentially connected to form a 1-pixel-thick polyline on the image grid. Subsequently, a binary morphological dilation with a disk-shaped structuring element of radius $r_0$ is applied, assigning a constant tubular thickness to the segments. 

To generate each binary mask, the number of segments $K$, the order $n$ of the Bézier curves, the displacement scale $\delta$ and the radius $r_0$ are all randomly sampled from an interval to ensure a wide variety of shapes. Table \ref{tab:vessshape_params} summarizes the parameters used in the VessShape dataset generation, along with their sampling ranges and descriptions.

\begin{table*}[t]
\caption{Main parameters used for generating the VessShape dataset.}
\label{tab:vessshape_params}
\centering
\begin{tabularx}{\textwidth}{l c X}
\hline
    \textbf{Parameter} & \textbf{Range} & \textbf{Description} \\
\hline
Number of curves $K$ & $[1,20]$ & Number of vessel segments generated per sample \\
Control points $n{+}1$ & $[2,20]$ & Controls the complexity of the Bézier curve \\
Displacement scale $\delta$ (px) & $[50,150]$ & Regulates the curvature/tortuosity of the vessel segments \\
Initial radius $r_{0}$ (px) & $[1,5]$ & Basal vessel radius before the smoothing operation \\
Matting blur $\sigma$ & $[1,2]$ & Standard deviation of the Gaussian used for blending the foreground and background \\
\hline
\end{tabularx}
\end{table*}

To compose the final image $I$ from a binary mask $M$, a foreground texture $F$ and a background texture $B$ are applied to the generated vessel segments and the background, respectively. The textures are randomly selected from the ImageNet dataset \cite{JiaDeng2009}. Specifically, for each mask $M$, two images are randomly drawn from two different classes of the ImageNet dataset. The images are then randomly cropped and resized to the target dimensions ($H \times W$). An alpha matte $A$ is then generated by smoothing $M$ with a Gaussian filter of standard deviation $\sigma$ and normalizing its values to the $[0, 1]$ range. The textures are subsequently blended using the following equation:

\begin{equation}
I \,=\, A\,F + (1-A)\,B,
\label{eq:compose}
\end{equation}
This blending operation ensures that vessel regions ($A \approx 1$) preserve the foreground texture while non-vessel regions ($A \approx 0$) retain the background texture. 

\begin{figure}[tbp]
    \centering
    \includegraphics[width=\columnwidth]{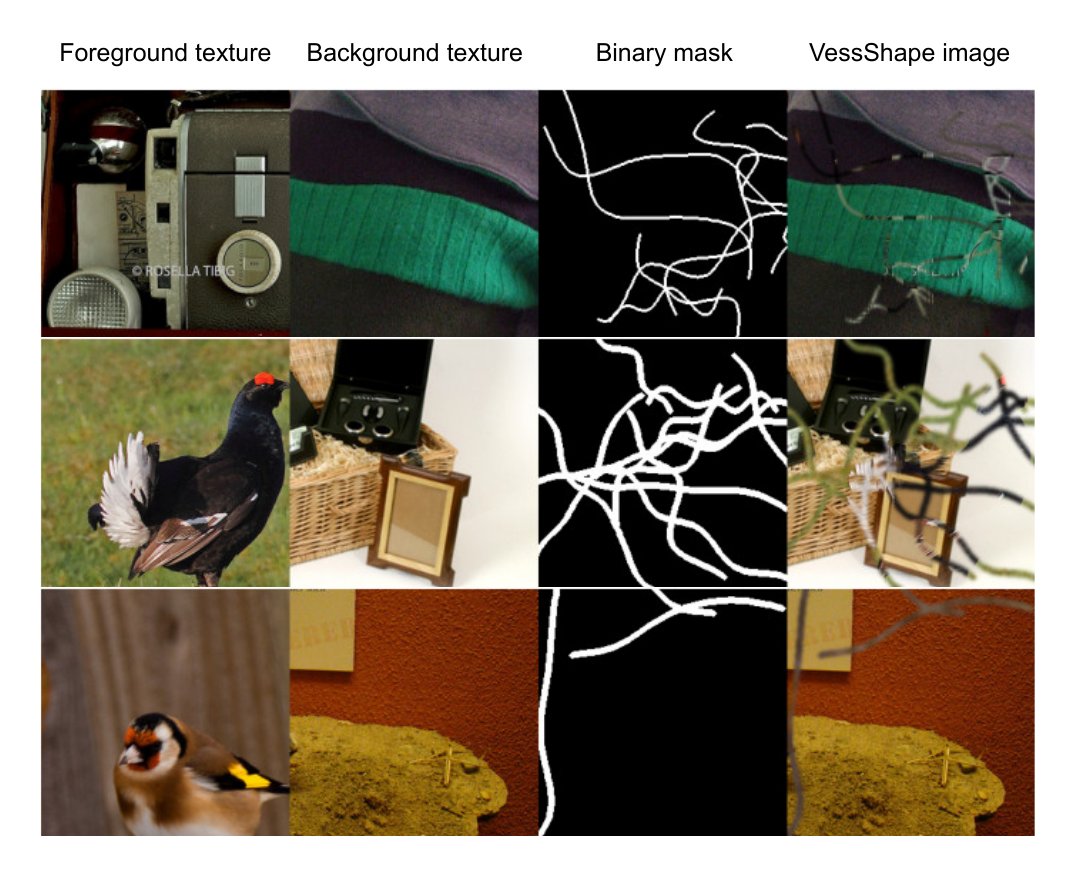}
    \caption{Examples of the VessShape generation process. Textures are sampled from ImageNet and blended according to procedurally generated binary masks to create the final VessShape images.}
    \label{f:vessshape_sample}
\end{figure}

The parameter $\sigma$ controls the smoothness of the vessel boundaries. Examples of generated masks and images are shown in Figure~\ref{f:vessshape_sample}.

\subsection{Real-world data for validation}

To quantify the usefulness of the shape bias introduced by VessShape, we consider two blood vessel datasets: DRIVE and VessMAP. The DRIVE dataset~\cite{Staal2004} serves as a popular standard for benchmarking retinal vessel segmentation algorithms and is composed of 40 fundus photographs split into 20 for training and 20 for testing, each measuring 584×565 pixels. In our experiments, all DRIVE images were converted to grayscale prior to training, validation, and testing. The VessMAP dataset~\cite{viana2025new} consists of 100 images, 256×256 pixels each, acquired by fluorescence microscopy of the mouse cortex. This dataset was curated to include a variety of challenging vascular characteristics, such as inconsistent noise and contrast levels, different vessel sizes, prominent imaging artifacts, and intensity fluctuations within vessel structures.

\begin{figure}[tbp]
    \centering
    \includegraphics[width=\columnwidth]{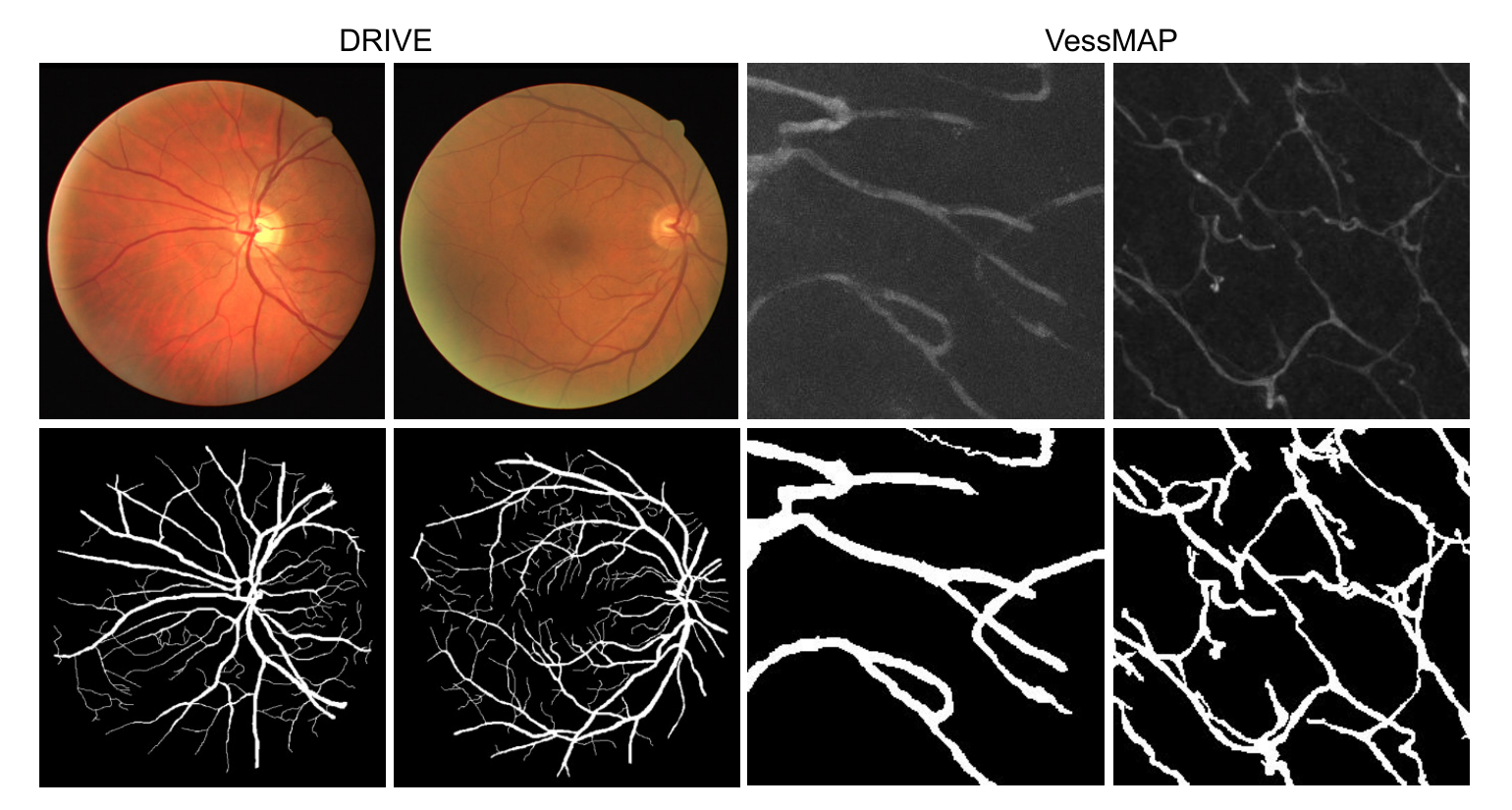}
    \caption{Samples from the DRIVE and VessMAP datasets and their respective ground-truth masks.}
    \label{f:drive_vessmap_samples}
\end{figure}

The two datasets originate from fundamentally different imaging modalities, resulting in distinct characteristics. The fundus images in DRIVE, which capture the entire retina, possess a clear global structure that includes landmarks like the optic disk. The samples also contain many very thin vessels which are challenging to segment. In contrast, the VessMAP images are highly magnified views of small cortical areas and have no discernible global organization. The borders of the vessels are generally less defined than those of the vessels in DRIVE. Another key difference is that, without any processing, the vessels in VessMAP are bright with dark backgrounds while the vessels in DRIVE are dark with bright backgrounds. Figure \ref{f:drive_vessmap_samples} shows samples from each dataset.

\subsection{Model architectures and training strategies}

We adopt a U-Net architecture with a symmetrical encoder-decoder design and skip connections between corresponding stages. Two models are compared, one with a ResNet18 encoder and the other with a ResNet50 encoder \cite{he2016deep}. The models were instantiated from the \textit{Segmentation Models Pytorch} Python package\footnote{\url{https://github.com/qubvel/segmentation_models.pytorch}}.

Two training scenarios are considered. In the first, training is done from scratch separately on the DRIVE and VessMAP datasets to establish a baseline. The second scenario consists of pre-training on the VessShape dataset and fine-tuning on DRIVE and VessMAP to measure the transferability and sample efficiency of the learned representations. These two scenarios involve three distinct training procedures: i) pre-training on VessShape, ii) fine-tuning on real-world data, and iii) training from scratch on real-world data. These procedures are described in the following subsections.

\subsection{Pre-training on VessShape}

The pre-training on VessShape aims to expose the model to a wide variety of tubular geometries while keeping the texture as a secondary cue. Each training sample is generated on the fly. This continual reshuffling of appearance paired with stable geometric rules should inject a strong shape bias while discouraging memorization of textures. The model is optimized to minimize the cross-entropy loss over this effectively infinite synthetic dataset. 

We pre-train two U-Net models with ResNet18 and ResNet50 encoders, referred to as VSUNet18 and VSUNet50. VSUNet18 was trained on approximately 7.1 million synthetic images over 8.6 hours, while VSUNet50 was trained on approximately 53.0 million images over 78.3 hours. Channel-wise normalization with ImageNet statistics was used for all input images. For evaluation during this stage, we used pre-generated validation and test sets from VessShape, containing 9,000 and 200 images, respectively. Table \ref{tab:vs_hparams} lists the hyperparameters used. Table~\ref{tab:vessshape_results_percent} summarizes the training performance on VessShape. All pre-training runs were executed on a workstation with 24 logical CPU cores and a single NVIDIA GeForce RTX 3090 GPU.

\begin{table}[t]
    \caption{Pre-training hyperparameters used for the VessShape dataset.}
    \label{tab:vs_hparams}
    \centering
    \begingroup
    \small
    \setlength{\tabcolsep}{6pt}
    \renewcommand{\arraystretch}{1.15}
    \begin{tabular}{l l l}
        \hline
        	\textbf{Hyperparameter} & \textbf{VSUNet50} & \textbf{VSUNet18} \\
        \hline
        Batch size & 96 & 192 \\
        Learning rate & $10^{-3}$ & $10^{-2}$ \\
        Weight decay & $10^{-4}$ & 0.0 \\
        \hline
    \end{tabular}
    \endgroup
\end{table}

\begin{table}[t]
    \caption{Performance of VSUNet variants after pre-training on the VessShape dataset. Values, presented in percentages, are the mean $\pm$ standard deviation evaluated on the fixed VessShape test set.}
    \label{tab:vessshape_results_percent}
    \centering
    \begingroup
    \small
    \setlength{\tabcolsep}{6pt}
    \renewcommand{\arraystretch}{1.15}
    \begin{tabular}{l r r}
        \hline
        	\textbf{Metric} & \textbf{VSUNet50} & \textbf{VSUNet18} \\
        \hline
        Dice & $86.1 \,\pm\, 2.2$ & $85.9 \,\pm\, 7.7$ \\
        Acc & $96.0 \,\pm\, 0.8$ & $95.6 \,\pm\, 3.7$ \\
        IoU & $75.8 \,\pm\, 3.2$ & $76.1 \,\pm\, 9.6$ \\
        Prec & $78.0 \,\pm\, 3.7$ & $77.4 \,\pm\, 9.6$ \\
        Rec & $96.4 \,\pm\, 1.2$ & $97.4 \,\pm\, 1.8$ \\
        \hline
    \end{tabular}
    \endgroup
\end{table}

\subsection{Fine-tuning}

We developed a systematic fine-tuning protocol\footnote{Code (VessShape pre-training + few-shot fine-tuning): \url{https://github.com/galvaowesley/vess-shape-experiments} } for few-shot training on 2D blood vessel datasets, which we apply here to DRIVE and VessMAP. The goal is to quantify performance gains as the number of labeled examples used for adaptation increases. For the VSUNet variants, we always start from the weights pre-trained on VessShape.

For each dataset $D$, we split the images into three disjoint subsets: $\mathcal{V}_{\text{train}}$, the pool of labeled images eligible for few-shot sampling; $\mathcal{V}_{\text{val}}$, used for auxiliary adjustments; and $\mathcal{V}_{\text{test}}$, held out for final evaluation. For DRIVE, we use 16 images for $\mathcal{V}_{\text{train}}$, 4 for $\mathcal{V}_{\text{val}}$, and 20 images for $\mathcal{V}_{\text{test}}$. For VessMAP, we adopt 60 images for $\mathcal{V}_{\text{train}}$, 20 for $\mathcal{V}_{\text{val}}$, and 20 for $\mathcal{V}_{\text{test}}$.

To apply fine-tuning with progressive sampling we define an ordered set of sample sizes $\mathcal{N} = \{ n_1, n_2, \ldots, n_K \}$ with $n_1 = 1$ and $n_K = n_{\mathcal{V}_{\text{train}}}$. For each $n \in \mathcal{N}$ we perform $R$ independent runs, and for each run $r$ we sample without replacement a training subset:
\[
\mathcal{V}^{(n,r)}_{\text{train}} = \text{sample}(\mathcal{V}_{\text{train}}, n)
\]

For each $n$, we keep the set of samples already used and sample previously unused samples. If there are no new samples (for large $n$ or high $R$), repetition is allowed. This procedure ensures that training runs are as diverse as possible.

Each subset $\mathcal{V}^{(n,r)}_{\text{train}}$ is used to fine-tune the model $S$ times ($s=1,\ldots,S$). The model is optimized to minimize the cross-entropy loss over $\mathcal{V}^{(n,r)}_{\text{train}}$, and performance is monitored on $\mathcal{V}_{\text{val}}$ after each epoch. The last checkpoint is then evaluated on $\mathcal{V}_{\text{test}}$. This approach allows for the decomposition of variance into: (i) training variability conditioned on a fixed image combination (within $(n,r)$) and (ii) variability across different image combinations between runs $r$.

In this work, we set $R=5$ and $S=3$. The sample size sequences $\mathcal{N}$ are $\{1,2,4,6,8,10, 12, 14, 16\}$ for DRIVE and $\{1,2,4,6,8,10, 12, 14, 16,18,20\}$ for VessMAP. We also consider the zero-shot case ($n=0$), in which the pre-trained model is evaluated directly on $\mathcal{V}_{\text{test}}$ without any adaptation on $D$.

\subsection{Training from scratch on real-world datasets}

We establish a baseline by training models directly on DRIVE and VessMAP without synthetic pre-training. We denote these models U-Net18 and U-Net50, and the training protocol is the same as the fine-tuning procedure. The only difference from the previous protocol is the lack of VessShape pre-training, which exposes the network to a noisier initial phase that is potentially more dependent on texture.

We do not define a zero-shot case for U-Net models because there is no useful prior state before observing at least one labeled image. The few-shot curves and the full-sample regime allow us to quantify: (i) the absolute gain provided by the shape bias acquired via VessShape; (ii) the difference in convergence speed as $n$ increases. In this way, the VSUNet versus U-Net comparison isolates the effect of shape bias while keeping all other factors controlled.

\section{Results}
\label{s:results}

\begin{figure*}[tbp]
    \centering
    \includegraphics[width=\textwidth]{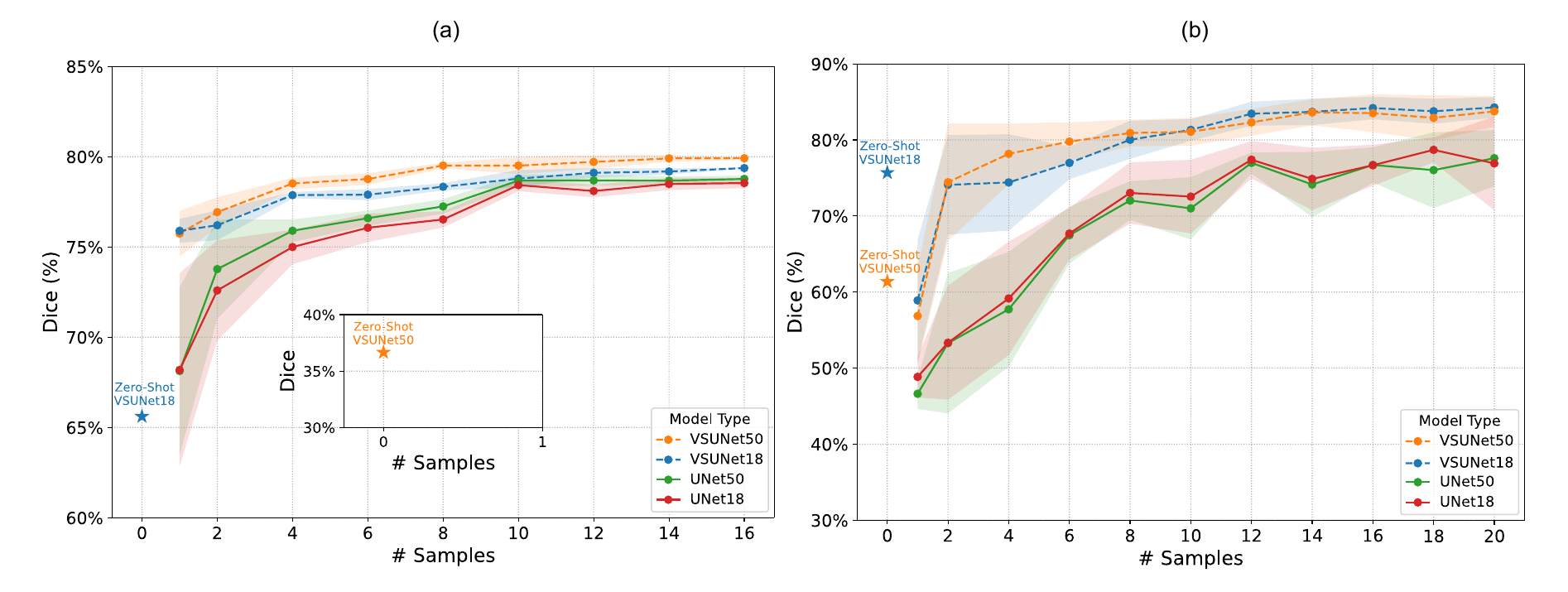}
    \caption{Few-shot and zero-shot Dice performance on (a) DRIVE and (b) VessMAP. Curves show the mean Dice over $R{=}5$ runs and $S{=}3$ repetitions for each sample size $n$. The shaded areas represent the standard deviation among runs. The inset shows the zero-shot Dice ($n{=}0$) for VSUNet50 on DRIVE, which was much lower than on the other experimental conditions.}
    \label{f:results_charts}
\end{figure*}

We conducted a quantitative analysis based on the models' performance curves as a function of the number of annotated examples. The main results are shown in Figure~\ref{f:results_charts}. We chose the Dice score as the main evaluation metric because it is widely used in medical segmentation tasks. Table \ref{tab:combined_fewshot_percent} summarizes the Dice scores, in addition to other key segmentation metrics, for zero-shot and few-shot segmentation over repeated runs.

\begin{table*}[t]
    \caption{Few-shot and zero-shot segmentation on VessMAP and DRIVE. Values, presented in percentages, are mean $\pm$ standard deviation over repeated training runs evaluated on the test set of each dataset. Zero-shot evaluations come from a single inference and thus have zero standard deviation.}
    \label{tab:combined_fewshot_percent}
    \centering
    \begingroup
    \small
    \setlength{\tabcolsep}{4pt}
    \renewcommand{\arraystretch}{1.15}
    \begin{tabular}{l c l l l l l l}
        \hline
        \textbf{Dataset} & \textbf{\#Examples} & \textbf{Model} & \textbf{Dice} & \textbf{Acc} & \textbf{IoU} & \textbf{Prec} & \textbf{Rec} \\
        \hline
        \multirow{10}{*}{DRIVE} & \multirow{2}{*}{0} & VSUNet18 & $\mathbf{65.6} \,\pm\, 0.0$ & $90.7 \,\pm\, 0.0$ & $49.0 \,\pm\, 0.0$ & $62.9 \,\pm\, 0.0$ & $69.9 \,\pm\, 0.0$ \\
         &  & VSUNet50 & $36.7 \,\pm\, 0.0$ & $88.8 \,\pm\, 0.0$ & $23.0 \,\pm\, 0.0$ & $72.8 \,\pm\, 0.0$ & $27.5 \,\pm\, 0.0$ \\
         \cline{2-8}
         & \multirow{4}{*}{1} & VSUNet18 & $\mathbf{75.9} \,\pm\, 0.7$ & $94.1 \,\pm\, 0.2$ & $61.2 \,\pm\, 0.9$ & $78.7 \,\pm\, 2.1$ & $74.1 \,\pm\, 1.9$ \\
         &  & VSUNet50 & $75.7 \,\pm\, 1.3$ & $93.9 \,\pm\, 0.6$ & $61.1 \,\pm\, 1.6$ & $77.3 \,\pm\, 4.6$ & $75.4 \,\pm\, 3.9$ \\
         &  & UNet18 & $68.2 \,\pm\, 5.4$ & $91.9 \,\pm\, 1.6$ & $52.3 \,\pm\, 5.8$ & $71.7 \,\pm\, 7.9$ & $69.0 \,\pm\, 11.8$ \\
         &  & UNet50 & $68.1 \,\pm\, 4.6$ & $91.6 \,\pm\, 3.1$ & $52.3 \,\pm\, 5.0$ & $72.2 \,\pm\, 9.9$ & $69.0 \,\pm\, 11.0$ \\
         \cline{2-8}
         & \multirow{4}{*}{16} & VSUNet18 & $79.4 \,\pm\, 0.0$ & $95.0 \,\pm\, 0.0$ & $65.8 \,\pm\, 0.0$ & $83.3 \,\pm\, 0.4$ & $76.2 \,\pm\, 0.3$ \\
         &  & VSUNet50 & $\mathbf{79.9} \,\pm\, 0.1$ & $95.2 \,\pm\, 0.0$ & $66.6 \,\pm\, 0.1$ & $84.6 \,\pm\, 0.3$ & $76.2 \,\pm\, 0.4$ \\
         &  & UNet18 & $78.5 \,\pm\, 0.3$ & $94.6 \,\pm\, 0.1$ & $64.7 \,\pm\, 0.4$ & $79.5 \,\pm\, 0.5$ & $78.1 \,\pm\, 0.4$ \\
         &  & UNet50 & $78.8 \,\pm\, 0.2$ & $94.7 \,\pm\, 0.1$ & $65.0 \,\pm\, 0.2$ & $80.7 \,\pm\, 0.6$ & $77.4 \,\pm\, 0.4$ \\
        \hline
        \multirow{10}{*}{VessMAP} & \multirow{2}{*}{0} & VSUNet18 & $\mathbf{75.7} \,\pm\, 0.0$ & $88.6 \,\pm\, 0.0$ & $61.6 \,\pm\, 0.0$ & $84.6 \,\pm\, 0.0$ & $69.6 \,\pm\, 0.0$ \\
         &  & VSUNet50 & $61.4 \,\pm\, 0.0$ & $81.7 \,\pm\, 0.0$ & $47.2 \,\pm\, 0.0$ & $74.6 \,\pm\, 0.0$ & $60.5 \,\pm\, 0.0$ \\
         \cline{2-8}
         & \multirow{4}{*}{1} & VSUNet18 & $\mathbf{58.9} \,\pm\, 8.0$ & $70.5 \,\pm\, 20.5$ & $45.5 \,\pm\, 8.8$ & $67.5 \,\pm\, 20.5$ & $73.8 \,\pm\, 14.8$ \\
         &  & VSUNet50 & $56.9 \,\pm\, 6.3$ & $70.8 \,\pm\, 20.3$ & $43.9 \,\pm\, 7.1$ & $68.3 \,\pm\, 20.8$ & $70.0 \,\pm\, 16.7$ \\
         &  & UNet18 & $48.8 \,\pm\, 2.7$ & $67.4 \,\pm\, 18.4$ & $36.2 \,\pm\, 3.2$ & $68.2 \,\pm\, 20.4$ & $62.2 \,\pm\, 21.0$ \\
         &  & UNet50 & $46.6 \,\pm\, 2.0$ & $66.5 \,\pm\, 18.1$ & $34.3 \,\pm\, 2.5$ & $67.4 \,\pm\, 20.6$ & $60.4 \,\pm\, 21.5$ \\
         \cline{2-8}
         & \multirow{4}{*}{20} & VSUNet18 & $\mathbf{84.3} \,\pm\, 1.3$ & $90.1 \,\pm\, 2.1$ & $73.9 \,\pm\, 1.8$ & $82.5 \,\pm\, 4.9$ & $88.8 \,\pm\, 4.8$ \\
         &  & VSUNet50 & $83.7 \,\pm\, 2.0$ & $90.5 \,\pm\, 2.5$ & $73.2 \,\pm\, 2.6$ & $85.1 \,\pm\, 3.2$ & $85.0 \,\pm\, 2.7$ \\
         &  & UNet18 & $76.9 \,\pm\, 6.2$ & $88.0 \,\pm\, 3.4$ & $64.5 \,\pm\, 7.5$ & $86.4 \,\pm\, 4.5$ & $73.7 \,\pm\, 9.4$ \\
         &  & UNet50 & $77.6 \,\pm\, 3.7$ & $88.0 \,\pm\, 3.5$ & $65.3 \,\pm\, 4.6$ & $85.5 \,\pm\, 4.1$ & $75.2 \,\pm\, 5.1$ \\
        \hline
    \end{tabular}
    \endgroup
\end{table*}

The performance curves in Figure \ref{f:results_charts} reveal behavioral differences in training between the VSUNet model variants and the models trained from scratch (U-Net). In both datasets, the VSUNet models start with a significant advantage in the few-shot regime, reaching a difference of 7 to 10 percentage points in the Dice score when training with a single sample. Furthermore, the curves of the pre-trained models rise more quickly and converge faster. In contrast, the curves of the U-Net models show a slower and more prolonged learning phase. This difference is also accentuated when analyzing the variance of the training runs, which tends to be lower for the VSUNet models. In the full-sample regime, when all samples are used, the Dice scores become more similar, but a performance gap between the models remains, indicating that the shape bias acquired from VessShape continues to offer benefits even with more labeled data available.

A counterintuitive phenomenon occurs in Figure \ref{f:results_charts}(b) for VessMAP, where the performance of the VSUNet models decreases when moving from zero-shot to one-shot, before recovering and surpassing the zero-shot performance with more samples. This behavior may be due to catastrophic forgetting \cite{MCCLOSKEY1989109}, where fine-tuning on a small dataset can lead the model to forget part of the previously acquired knowledge. However, as more data are introduced, the model can recover and even improve its performance, suggesting that the initially learned shape bias is robust and can be reinforced with additional data.

The one-shot decrease in accuracy has also been documented in large-scale models such as CLIP \cite{Radford2021LearningTV}. The CLIP model has strong zero-shot capability because it has learned prior rules that can be accessed with prompts such as ``a photo of a \{label\}''. In our case, the VessShape pre-training induces a strong shape prior that functions similarly to this explicit knowledge. In contrast, when fine-tuning is performed with a single sample, the model is forced to optimize its weights based on very limited information, which contains not only the desired shape but also instance-specific details, such as noise and texture. This leads to overfitting, as instead of learning the relevant features of the new domain, the model tends to memorize specific details of that single example, such as its particular noise and texture pattern.

However, this phenomenon is not observed in the DRIVE dataset. This suggests that the initial shape prior is not as well-aligned with the DRIVE domain. Thus, the first training sample provides new and valuable information that helps the model adjust its weights in a beneficial way, rather than causing overfitting. Similarly, the model also benefits from more examples, as demonstrated in Figure \ref{f:results_charts}(a). One reason for the lower performance might be that the blurring applied to VessShape samples tends to generate thicker vessels than those observed in the DRIVE dataset. Tuning the parameters to better reflect the characteristics of retinal vessels could lead to better results. However, the VessShape samples used in the experiments were created without considering specific datasets since the objective is to evaluate the potential of using generic shape priors for pre-training.

\begin{figure*}[tbp]
    \centering
    \includegraphics[width=\textwidth]{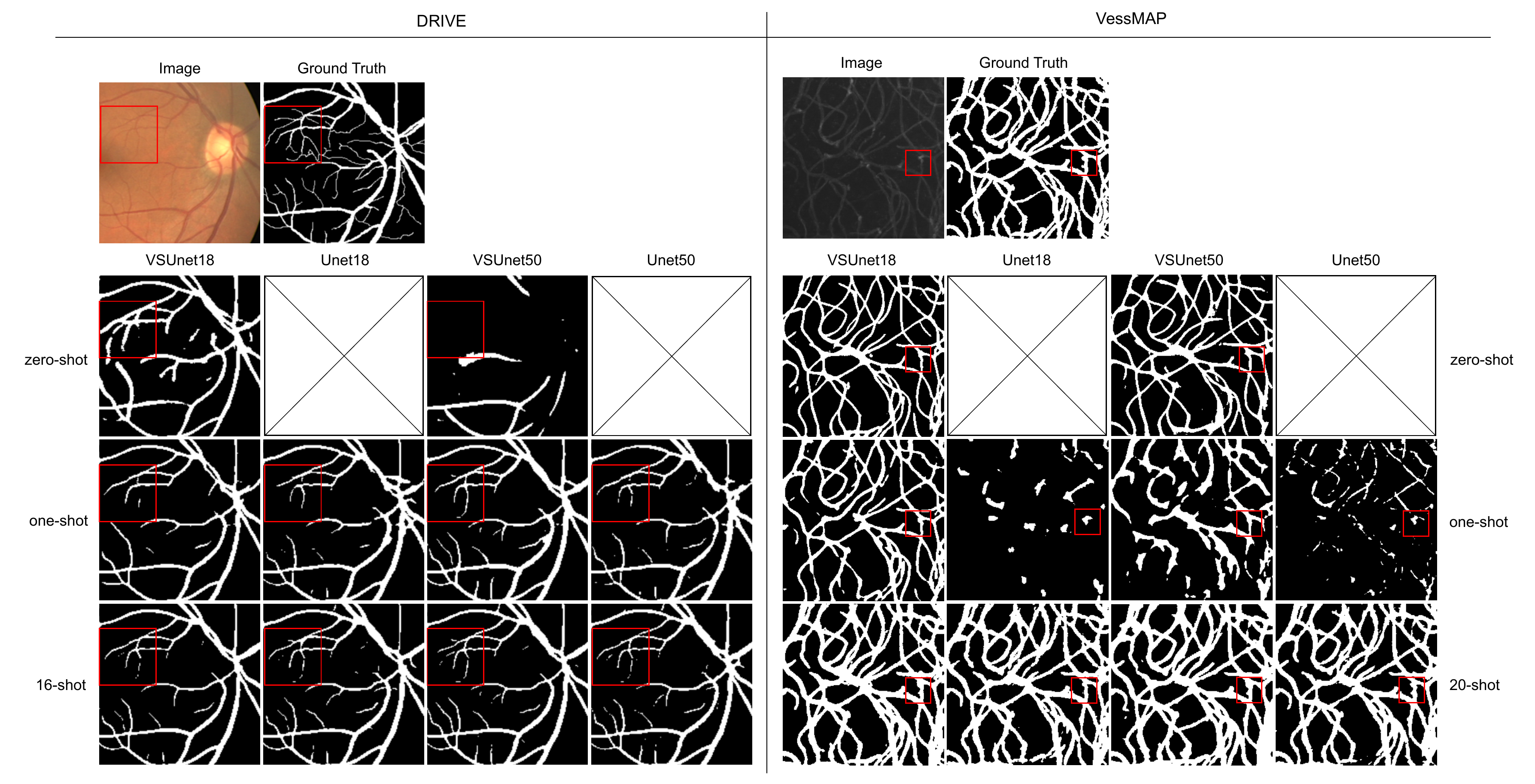}
    \caption{Qualitative visual comparison of segmentation performance on the DRIVE and VessMAP datasets. The figure shows the outputs of VSUNet and U-Net variants across different few-shot regimes: zero-shot, one-shot, and few-shot (16-shot for DRIVE and 20-shot for VessMAP). For the U-Net variants, no zero-shot output is available. Red squares highlight specific regions of interest to facilitate comparison: an area with low-caliber vessels in DRIVE and a vascular bulge in VessMAP.}
    \label{f:results_fewshots_drive}
\end{figure*}

Our analysis is complemented by a qualitative evaluation of the generated segmentations. Figure~\ref{f:results_fewshots_drive} illustrates the differences between the results of the models and the impact of the shape bias. For the DRIVE dataset, the VSUNet18 zero-shot model is able to correctly segment thick vessels and successfully delineate the main vascular structure, while VSUNet50 shows many false negatives. With just one sample, both VSUNet models adapt quickly, and the VSUNet50 now exhibits superior segmentation of fine vessels, as highlighted by the red square. The U-Net variants were also able to segment the main structure and the region of interest with reasonable quality in the evaluated example. With 16 samples, the visual differences between all models become minimal, converging to similar results.

For VessMAP, the highlighted red region in the figure shows a vessel bulge with a sharp change in intensity. The zero-shot VSUNet models capture the general topology, but not the shape of the bulge. The one-shot scenario confirms the performance drop, with the VSUNets generating discontinuous results and the U-Nets, without prior knowledge, producing very low-quality segmentation maps. Still, the VSUNet18 model shows an initial refinement of the bulge geometry. With 20 samples, all models improve significantly, although the U-Net variants still exhibit minor continuity flaws compared to the more cohesive outputs of the VSUNet models.

Despite being smaller and trained with significantly less data than VSUNet50, the VSUNet18 model showed superior zero-shot performance and remained competitive across all experiments. Thus, the model offers an interesting balance between efficiency and generalizability.

One of the most significant results is the generalization capability of the models, demonstrated by their zero-shot performance. The models can segment the vessels in visually dissimilar domains without any specific training. In VessMAP, the vessels have high intensity on a dark background, whereas in DRIVE, the vessels are dark on a light background. The VSUNet18 model can successfully segment a large number of vessels in both domains without fine-tuning. This generalization ability suggests that the model has learned a robust representation of vessel shapes, which is applicable regardless of the specific visual characteristics of the domain.

\section{Conclusion}
\label{s:conclusion}

We proposed a strategy focused on instilling a strong shape bias in deep learning models by pre-training them on VessShape, a large-scale synthetic dataset generator. By combining simple, universal vessel-like geometries with highly diverse textures, VessShape encourages models to learn robust vessel shape priors instead of domain-specific texture features.

Our experiments demonstrated the effectiveness of this approach. A model pre-trained on VessShape achieved strong few-shot performance on two distinct and challenging real-world datasets, DRIVE (retinal fundus photography) and VessMAP (cerebral cortex microscopy), requiring only a few annotated samples to adapt to these new domains. Furthermore, the model exhibited remarkable zero-shot capabilities, successfully identifying vessel structures without any fine-tuning. These results confirm our hypothesis that leveraging a geometric prior is an effective strategy to improve data efficiency and model robustness against domain shifts.

While our current work focused on generic priors, the VessShape framework is flexible enough to be tuned to the characteristics of a specific vasculature. For instance, its parameters could be adjusted to generate the thinner, sharper vessels typical of retinal images, in contrast to those in VessMAP. We consider such domain-specific optimization a promising direction for future work, as our primary objective here was to evaluate the impact of general shape priors.

There are many avenues for future work. The current geometry generation in VessShape, based on Bézier curves, could be extended to include more complex and biologically plausible vascular topologies, such as true bifurcations and network-like structures. Furthermore, extending the VessShape generation framework to 3D would allow this pre-training strategy to be applied to volumetric medical imaging modalities such as CT and MRI. Future studies could also explore the application of a shape-centric pre-training approach to the segmentation of other tubular structures in biology, such as neurons or airways.

Our work underscores an interesting principle: for segmentation tasks with strong and consistent shape priors, focusing on core geometric features can be a more effective and generalizable pre-training strategy than attempting to mimic the appearance of a single target domain.

\section*{Ethics statement}

This study was conducted using publicly available, anonymized datasets and did not involve any new experimentation with human or animal subjects. Therefore, in accordance with institutional and national guidelines, this research was exempt from review by an ethics committee, and the requirement for informed consent is not applicable.

\section*{Acknowledgements}
This study was financed, in part, by the São Paulo Research Foundation (FAPESP), Brasil. Process Number \#2025/04800-9.

\bibliography{references}

\end{document}